\documentclass[conference]{IEEEtran}
\IEEEoverridecommandlockouts
\usepackage{cite}
\usepackage{amsmath,amssymb,amsfonts}
\usepackage{algorithmic}
\usepackage{graphicx}
\usepackage{textcomp}
\usepackage{xcolor}

\usepackage{multirow}

\def\BibTeX{{\rm B\kern-.05em{\sc i\kern-.025em b}\kern-.08em
    T\kern-.1667em\lower.7ex\hbox{E}\kern-.125emX}}
\begin{document}

\title{Quadruplet Loss For Improving the Robustness to Face Morphing Attacks \\
}

% \and
% \and
% \and
% \and
% \and

\author{Iurii Medvedev\\
\textit{$^1$Institute of Systems}\\
\textit{and Robotics,}\\
\textit{University of Coimbra,}\\
Coimbra, Portugal\\
{\tt\small iurii.medvedev@isr.uc.pt}
\and
Nuno Gonçalves $^{1,2}$\\
\textit{$^2$Portuguese Mint and Official}\\
\textit{Printing Office (INCM),}\\
\textit{Lisbon, Portugal}\\
{\tt\small nunogon@deec.uc.pt}
}

\maketitle

\begin{abstract}
Recent advancements in deep learning have revolutionized technology and security measures, necessitating robust identification methods. Biometric approaches, leveraging personalized characteristics, offer a promising solution. However, Face Recognition Systems are vulnerable to sophisticated attacks, notably face morphing techniques, enabling the creation of fraudulent documents.
In this study, we introduce a novel quadruplet loss function for increasing the robustness of face recognition systems against morphing attacks. Our approach involves specific sampling of face image quadruplets, combined with face morphs, for network training. Experimental results demonstrate the efficiency of our strategy in improving the robustness of face recognition networks against morphing attacks. 
\end{abstract}

\begin{IEEEkeywords}
quadruplet loss, face morphing, face recognition, computer vision, deep learning
\end{IEEEkeywords}

\section{Introduction}
%intro
In recent years, the evolution of deep learning technologies has profoundly influenced the technology and modern security measures. Security concerns continue to expand prompting the need for more robust and reliable identification methods. To address these challenges, biometric approaches have emerged as a promising solution, personalized characteristics for the recognition process.

Among the various modalities of biometric identification, face recognition has garnered significant attention and adoption due to its simplicity of acquisition and recent advancements in computer vision techniques. The widespread use of Face Recognition Systems (FRSs) underscores the importance of facial traits in modern biometric applications, facilitating identification and verification processes across various domains. However,  FRSs remain vulnerable to attacks, particularly from sophisticated image manipulation techniques that aim to deceive the system.

One such technique is face morphing, which involves merging or blending two or more digital face images to create a synthetic image that shares biometric properties with the originals, potentially matching different individuals. Face morphing can facilitate acceptance of manipulated face images for the creation of fraudulent documents, which can then be utilized by unauthorized individuals engaging in fraudulent activities. While such fraudulent documents are occasionally detected during border control procedures, the true extent of the existence of fraudulent documents remains unknown \cite{torkar2023morphing}.
The emergence of deep learning techniques has significantly advanced the field of face recognition, yet face morphing poses a persistent security risk, challenging traditional human or computer-based recognition methods. Ensuring the robustness of face recognition systems against face morphing attacks is crucial for maintaining the integrity of biometric identification. 

In this work, we follow and extend contrastive methods in face recognition and introduce a novel quadruplet loss function aimed at enhancing the robustness of face recognition systems to morphing attacks. Our approach involves the specific sampling of quadruplets of face images, combined with face morphs, to train the network. Results of our experiments demonstrate that our strategy effectively increases the robustness of face recognition networks against morphing attacks, thereby contributing to the advancement of biometric security measures.

%With a particular emphasis on the importance of facial modalities in biometric systems, this paper aims to contribute to the ongoing efforts to enhance the security and reliability of biometric identification technologies.

%face recognition

%face morphing and risks

%face morphing robustness

\section{Related Work}

In this section, we will review recent advancements in face morphing and face recognition that are relevant to our research objectives.

\subsection{Face Morphing}

Generation of a face morph from original images usually includes several stages: \textit{extracting facial features} $\rightarrow$ \textit{averaging features} $\rightarrow$ \textit{generating a morphed image from features} $\rightarrow$ \textit{image refining/retouching}. Early methods \cite {magic_passport, ubo_morpher} relied on facial landmark features for aligning original images and generating face morphs through image blending. However, many of these techniques are prone to producing images with blending artifacts.

Recent advancements in generative deep learning methods have led to the development of face morphing methods that utilize the deep latent feature domain. These approaches may employ a range of deep learning tools, such as variational autoencoders (VAE)  \cite{morGAN} or generative adversarial networks (GANs)  \cite{Can_GAN_Morphs,MIPGAN_morphing_paper} or diffusion autoencoders \cite{MorDIFF}.

%Mention face morphing detection
Significant number of recent research studies has been dedicated to the detection of face morphs within biometric systems.  \cite{morphing_detection_print-scan,morphing_fusion, MorDeephy,neto2022orthomad}.

%bridge increasing robustness
At the same time the issue of the robustness and stability of facial template (representation) is also very important \cite{Biometric_Systems_under_Morphing_Attacks}, and even famous private benchmarks begin to support performance estimation for this  \cite{bench_NIST_morph}.

%several methods for increasing robustness
For instance Marriott et al.\cite{Marriott2020RobustnessOF}  tested several  facial recognition algorithms to the robustness against face-morphing attacks and demonstrated the potential threat of face morphing attacks.

\subsection{Face Recognition}% via Contrastive Losses}
%Intro
Deep learning tools, known for their effectiveness in pattern recognition tasks, have found widespread application in various biometric fields, including face recognition \cite{ImageNet_cite}. Among these tools, convolutional neural networks (CNNs) have gained prominence for their capacity to extract discriminative features from unconstrained images, making them particularly efficient in this context.

%Strategies
Recent approaches in face recognition primarily focused on extracting a compact facial biometric template from deep features generated by a backbone network. The main objective is to enhance the discriminative capability of this template under specific conditions. The strategies for training deep networks in face recognition can be categorized into classification and contrastive approaches.

%classification approaches
Many recent methods or training face recognition networks approach it via a multi-class closed-set classification problem using existing face image datasets. The discriminative information learned by the network is embedded in its hidden feature layer, which can be utilized for open-set identity discrimination tasks. These methods typically employ softmax loss and its variations for classification \cite{deepid2_plus_paper}. 
%marginal
Softmax classification-based methods were broadly modified by techniques to enhance intra-class compactness and inter-class separation. \cite{centerface_paper, sphereface_paper, cosface_paper, arcface_paper, addictive_margin_paper, equalized_margin_paper} and with sample-specific strategies for additional  control of the feature domain with samples hardness, classification uncertainty or quality \cite{npcface, probabilistic_embedding, Magface, QualFace, QualFace2}.

%metric learning
Contrastive methods (or metric learning methods) utilize the target similarity metric (for instance euclidean or cosine distance) to straightforwardly optimize the distance between deep features by matching face image pairs during the learning process \cite{chopra_metric_paper}. 
These methods are usually characterized by the high demands of dataset diversity and sophisticated sampling strategies for reliable convergence of the training process \cite{working_hard}. They are shown to be effective on large datasets and useful in the tasks of transfer learning and fine-tuning.

%hu et al
Initial works straightforwardly optimised pairwise distance contrast utilising small neural networks for feature extraction \cite{Nguyen2010CosineSM, Hu2014DiscriminativeDM}.

%Facenet
The major improvement of contrastive methods in face recognition was introduced by the FaceNet approach \cite{facenet}. FaceNet proposed the use of triplet loss, which directly optimizes the embedding of face images in a high-dimensional space by the distances within triplets of samples. Triplets are composed of an anchor image, a positive image of the same identity as the anchor, and a negative image of a different identity. This approach required sophisticated sampling of those triplets to learns discriminative feature representation that ensures close proximity between embeddings of images of a same individual while maximizing the separation between embeddings of images of different individuals. Many of the following works on contrastive learning methods in face recognition were inspired by the triplet loss strategy. 

%Docface
For instance DocFaceID \cite{DocFaceID} introduces the MPS loss function for fine-tuning a pair of sibling networks for the ID document photo matching
problem. MPS loss aims to enhance representation learning by maximizing the margin between match pair similarities and non-match pair similarities. This approach simulates a scenario where ID photos serve as templates, while selfies from different subjects act as probes seeking verification, or vice versa. 

%CoReFace
CoReFace \cite{coreface} revisited the contrastive learning approach for face recognition by incorporating an adaptive margin for better training regularisation. Additionally, it proposed a new pair-coupling protocol to address the similarity issue stemming from pair symmetry, thereby enhancing the effectiveness of the approach.

\subsection{Quadruplet loss}
%intro
As an evolution of traditional triplet loss, the Quadruplet loss has emerged as a compelling method in the realm of deep metric learning for face recognition. By leveraging four samples instead of three, the Quadruplet loss can offer enhanced discriminative power and robustness. However such approach can increase the complexity of efficient sampling strategy. 

%reidentification
The Quadruplet loss found its application in the tasks of Re-Identification. For instance, it was applied to the Re-Identification of baggage \cite{QuadNet} or persons \cite{quadruplet_person_reid, Chen2017BeyondTL}.

%HFR
The Quadruplet loss was also used in applications for  Heterogeneous Face Recognition where it addresses challenges posed by cross-domain discrepancies and limited training data \cite{HFR}. It integrates domain-level and class-level alignment within a unified network to learn domain-invariant discriminative features (DIDF). Domain-level alignment reduces distribution discrepancy, while a specialized quadruplet loss further diminishes intra-class variations and enhances inter-class separability. %By generating cross-domain image quadruplets and imposing explicit constraints on feature distances, the framework effectively mitigates class misalignment. 

%object tracking
In applications of visual tracking the Quadruplet loss was used to develop a discriminative model to one-shot learning (recognize objects of the same class by a single exemplar given) and helped to improve tracking accuracy and real-time speed \cite{Dong2017QuadrupletNW}.

\section{Methodology}

In our work we intend to leverage the potential of quadruplet loss by attaching additional branch which inputs the face morph sample to the common triplet.

\subsection{Quadruplet Loss With Face Morphing}

%One of the most important papers here is published by \cite{facenet}. The essence of the method is the learning and optimizing of the face embedding itself with using a triplet loss function. 

In order to define the Quadruplet loss in our work we revising the formulation of the triplet loss in the FaceNet approach \cite{facenet}.
The triplet loss is computed for three face images (anchor $x_{i}^{a}$, positive $x_{i}^{p}$ and negative $x_{i}^{n}$) where anchor and a positive belong to the same identity and negative belong to another identity. Next the distance between anchor and a positive is minimised and the distance between anchor and a negative is maximised:

\begin{equation}
\label{eq:triplet_loss}
\left \| x_{i}^{a} - x_{i}^{p} \right \|_{2}^{2} +\alpha < \left \| x_{i}^{a} - x_{i}^{n} \right \|_{2}^{2}, \forall  (x_{i}^{a},x_{i}^{p},x_{i}^{n}) \in \tau 
\end{equation}
where $\alpha$ is a margin that is enforced between positive and negative pairs. $\tau$ is the set of all possible triplets in the training set.

The embedding is represented by $f(x) \in \mathbb{R}^d$  and the loss that is being minimized is then:

\begin{equation}
\label{eq:triplet_loss_loss}
L =\sum (\left \| f(x_{i}^{a}) - f(x_{i}^{p}) \right \|_{2}^{2} -\left \| f(x_{i}^{a}) - f(x_{i}^{n}) \right \|_{2}^{2} +\alpha)
\end{equation}

%A very important part is the choice of the images triplets because while generating all possible ones many of them already easily satisfy the loss function. \cite{facenet} achieve that by selecting the hard positive/negative exemplars from within a mini-batch. However at the same time choice of triplet loss leads to several problems. First there are a number of convergence issues reported, that is why careful choice of selection strategy is crucial.  Also the samples selection leads to the requirements of relatively much larger datasets in comparison with classification approaches.

\begin{figure}[htbp]
\centerline{\includegraphics[width=0.9\linewidth]{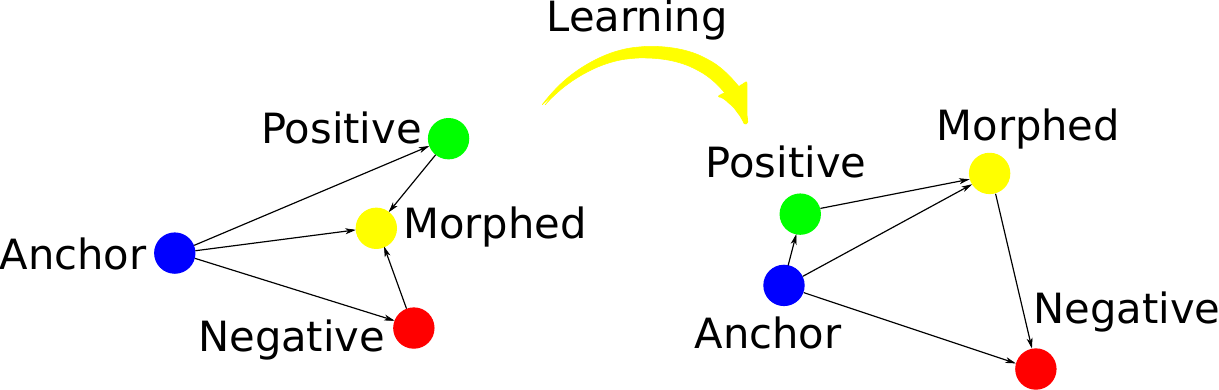}}
\caption{The proposed Quadruplet Loss minimizes the distance between an anchor and a positive, both of which have the same identity, and maximizes the distance between the anchor and a negative, anchor and morphed, negative and morphed where anchor and negative belongs to different identities and morphed is taken between the positive and negative.}%Quadruplet loss minimizes the \textit{anchor-positive} distance and maximizes the \textit{positive-negative}, \textit{anchor-negative}, \textit{positive-morph}, \textit{morph-negative}.}
\label{quadruplet_loss}
\end{figure}

We modify this formulation by inserting a morph sample into the consideration. This face morph sample may be generated from two face images, which belong to the same identities of positive  samples. The source images for generating may be indeed different from other components of the initial triplet. With this modification the cross-sample distance condition have the following formulation:

%In fact the formulation of quadruplet loss is not novel and for instance was used in person (not face) recognition by  We intend to revisit its formulation for the purposes of face recognition with morphing attacks:

\begin{equation}
\begin{aligned}
\label{eq:quadruplet_loss}
\left \| x_{i}^{a} - x_{i}^{p} \right \|_{2}^{2} +\alpha < \omega _{a/n}\left \| x_{i}^{a} - x_{i}^{n} \right \|_{2}^{2}+ \\ +\omega _{a/m}\left \| x_{i}^{a} - x_{i}^{m} \right \|_{2}^{2}+ \omega _{n/m}\left \| x_{i}^{n} - x_{i}^{m} \right \|_{2}^{2}+\\ +\omega _{p/m}\left \| x_{i}^{p} - x_{i}^{m} \right \|_{2}^{2}, \forall  (x_{i}^{a},x_{i}^{p},x_{i}^{n},x_{i}^{m}) \in \mathbb{Q}
\end{aligned}
\end{equation}
where $x_{i}^{a},x_{i}^{p},x_{i}^{n},x_{i}^{m}$ correspond to anchor, positive, negative, and morphed (for positive/negative pairs) in the quadruplet $\mathbb{Q}$ and $\omega _{i}$ signify the weight for each part of the sum, which are selected to balance identity matched and non-matched distances ($\omega _{i} = 0.25$) on Fig.~\ref{quadruplet_loss}.

\begin{figure}[htbp]
\centerline{\includegraphics[width=0.9\linewidth]{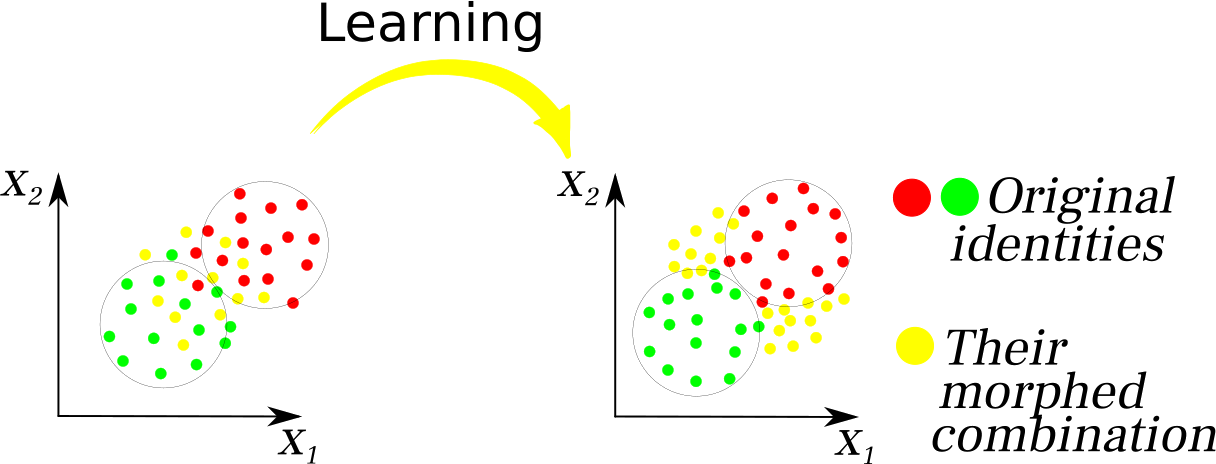}}
\caption{The distribution of features $x$ (with two dimensional feature layer) with common face recognition approaches (left) and desired feature distribution in proposed approach (right).  Each point on 2D surface corresponds to a single image features (data is not based on real experiments).}%Desired transformation of feature distribution}
\label{feature_distribution}
\end{figure}

Such methods in fact is directed onto achieving the maximum discriminative power of the learned features, while also enlarging variances between original images and their morphed combinations. 
The approximate visualisation of desired feature distribution (in case of 2D feature layer dimensionality) is depicted on the Fig.~\ref{feature_distribution}.

\subsection{Data Harvesting}
Our training data is based on the collection, which is developped and used in \cite{MorDeephy}. It is build on the filtered version of VGGFace2 dataset\cite{VGGface2}, where quality-based filtering is used to select a subset of images suitable for face morphing (frontal image with acceptable quality). Morphs are generated from those filtered originals with  a customized landmark-based morphing approach is utilized with a blending coefficient set to 0.5. Additionally, GAN-based morphs are generated using the StyleGAN method. These morphs are synthesized by projecting original images onto a latent domain, followed by interpolation of their deep representations to produce the resulting morph.
We also employ "selfmorphs" generated by applying face morphing to images of the same identity. These selfmorphs serve as bona fide samples, allowing to focus on the behavior of deep face features while mitigating the influence of perceptual artifacts. In our work we assume that deep discriminative face features remain unaffected by selfmorphing.

\subsection{Data Sampling}
\begin{figure}[htbp]
\centerline{\includegraphics[width=0.7\linewidth]{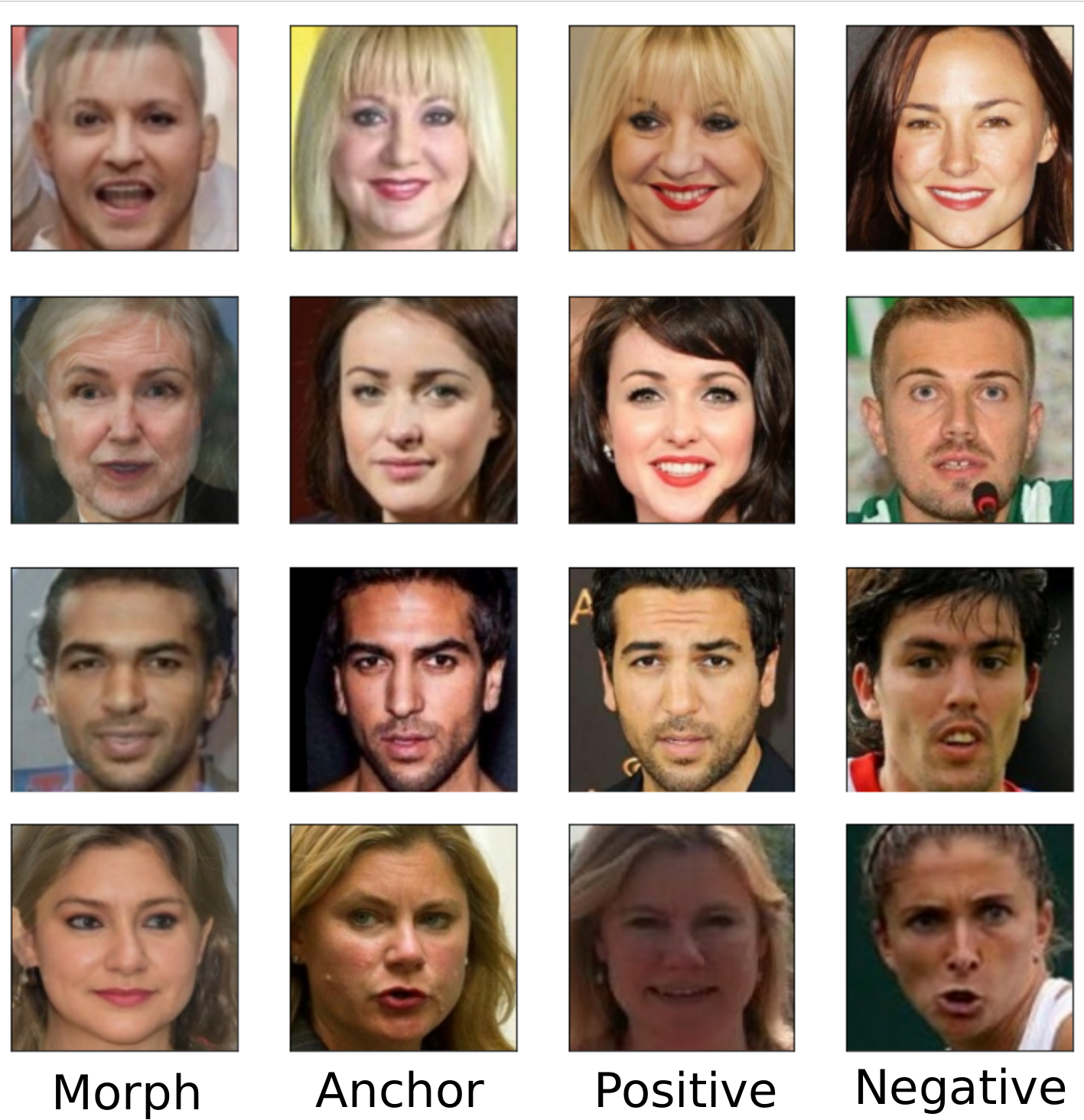}}
\caption{Quadruplet samples}
\label{quadruplet_samples}
\end{figure}

%sampling
Training data generator combines the batches of quadruplets with the above data. In our sampling strategy we go thought the list of available face morphs and for each one we select anchor, positive and negative from the source morph sample identities to combine the quadruplet (see Fig.~\ref{quadruplet_samples}). 
%Add here a quadruplet
The resulting quadruplets are combined into batches and fed to the Siamese network input.

\begin{figure*}[htbp]
\centering

\includegraphics[width=0.42\linewidth]{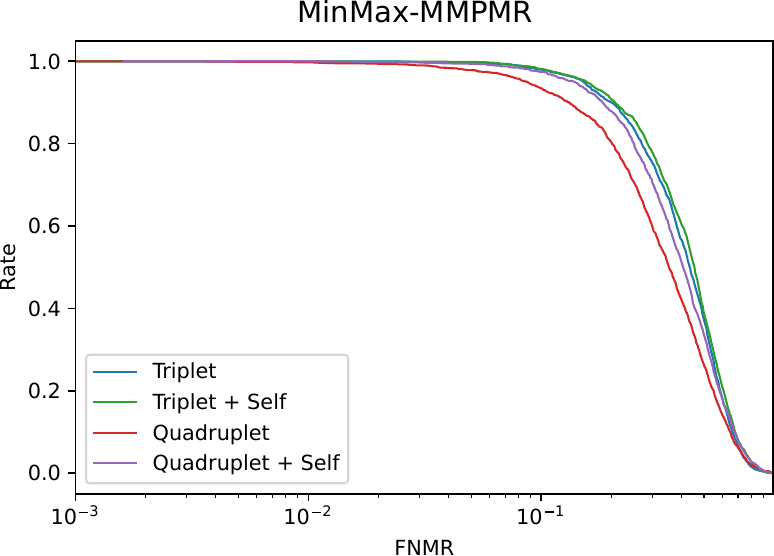}
\includegraphics[width=0.42\linewidth]{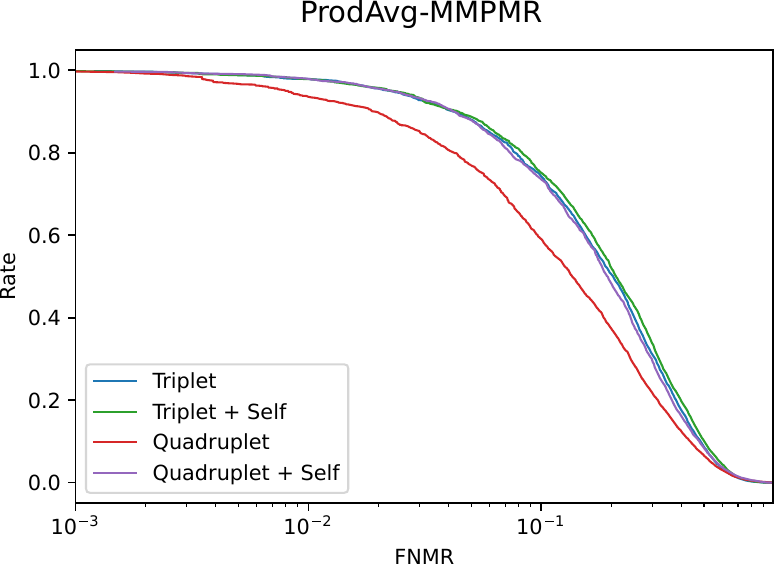}

 a \hspace{220pt}   b

\vspace{5pt}
\includegraphics[width=0.42\linewidth]{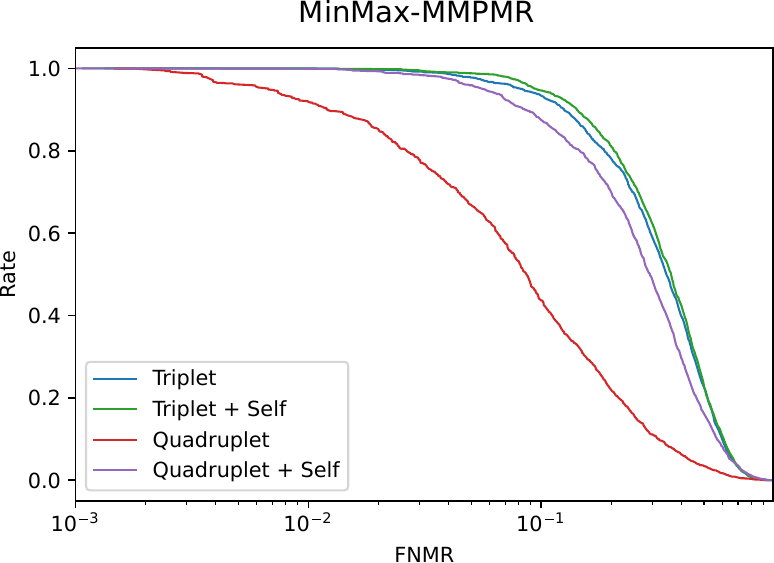}
\includegraphics[width=0.42\linewidth]{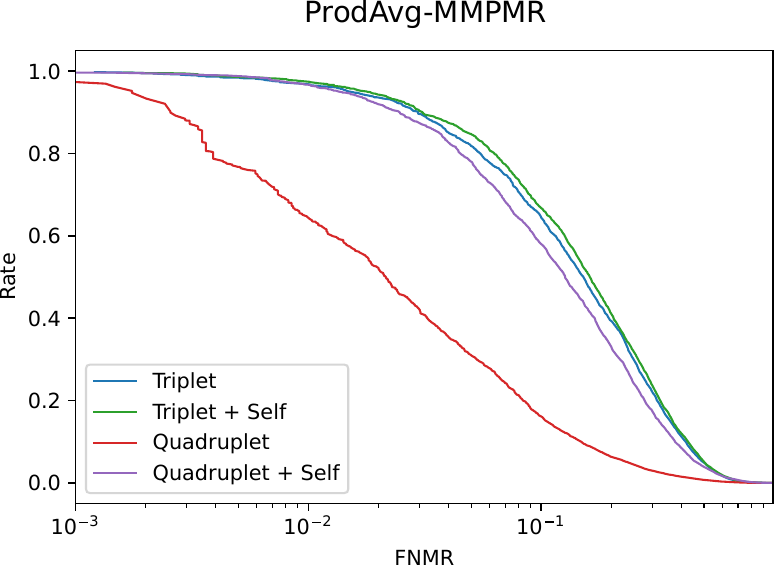}

 c \hspace{220pt}   d

\caption{MinMax-MMPMR and ProdAvg-MMPMR for different models and protocols as a dependency from FNMR. a,b - LDM benchmark; c,d - STG benchmark.}
\label{fig:MMPMR_1}
\end{figure*}

\section{Experiments}
We perform several experiment with our methodology and selected data.
In all our experiments we train the Resnet-50 architecture with image size $300\times300$ pixels by 10 epochs with Adam optimizer. However first we need to discuss the metrics and benchmarks that will be used for performance estimation.

\subsection{Benchmarking}

%explain 2 metrics
There are several metrics for estimating the robustness of face recognition system to face morphing. Following the \cite{Biometric_Systems_under_Morphing_Attacks} we compute the MinMax-MMPMR (Eq. \ref{eq:minmax}) and ProdAvg-MMPMR (Eq. \ref{eq:prod}), which are motivated by the international standard ISO/IEC 30107-3 \cite{REFERENCE_OF_ISO_30107}.

\begin{equation}
\label{eq:minmax}
\begin{aligned}
MinMax-MMPMR(\tau ) = {\color{white} \cdot\cdot\cdot\cdot\cdot\cdot\cdot\cdot } \\ \frac{1}{M} \cdot \sum_{m=1}^{M} \left \{ \left ( \begin{matrix}
min\\ 
i =1,...,N_{m}
\end{matrix} \left [\begin{matrix}
max\\ 
i =1,...,I_{m}^{n} 
\end{matrix}S_{m}^{n,i}  \right ]\right )>\tau \right \}
\end{aligned}
\end{equation}
where $I_{m}^{n} $ is the number of samples of subject $n$ within morph $m$. %MinMax-MMPMR also models the case where $N_m$ subjects launch single attacks to several biometric authentication systems ($I_{m}^{n}= 1$).

\begin{equation}
\label{eq:prod}
\begin{aligned}
ProdAvg-MMPMR(\tau ) = {\color{white} \cdot\cdot\cdot} \\ \frac{1}{M}\cdot \sum_{m=1}^{M} \left \{ \prod_{n=1}^{N_{m}} \frac{1}{I_{m}^{n}} \cdot \sum_{i=1}^{I_{m}^{n}}  \left \{ S_{m}^{n,i}  >\tau  \right \} \right \}
\end{aligned}
\end{equation}

%explain carlas work
To estimate those metrics we build a custom benchmark basing on the dataset from \cite{GuerraMG23}. This dataset is collected fur the studies of ICAO-requirements\cite{ICAO_requirements_1} conformity and thus contain both ICAO-compliant and non-compliant images. One of the benefits of the data is that it contain images collected with a professional digital camera and a smartphone giving these two modalities of the data.
%explain separation to enrollment and reference
We use those different modalities to simulate the differences of enrollment and life capture (reference) in our benchmark utilities. Namely we select \textit{enrollment} images from the ones that are collected with professional digital camera and the \textit{reference} from the smartphone images.  This also imply strict ICAO-compliance for the enrollment images and relaxed conditions for the reference ones. In total we have 1567 and  8760 images correspondingly.

%explain fnmr
In order to normalize the result metric rates they are usually computed at the same level of FNMR in some standard face recognition 1-1 performance value. We define such standard 1-1 verification protocol by pairing \textit{enrollment} and \textit{reference} images. This protocols consist of $\sim$71k of pairs ($\sim$2k match pairs and $\sim$69k non-match pairs) 

%protocols
We define a standard face pairing protocol for morphing, which standardize the images selection for performing face morphing. This protocol consist of 2142 pairs, thus giving the same amount of face morphs in our benchmarks. face morphs are generated only from the \textit{enrollment} images.
%generated morphs
Following this protocols we generate face morphs with a landmark based approach (LDM) and a StyleGAN (STG)\cite{styleGAN}  approach to have in total two different benchmarks for performing comparisons in our work.

\subsection{Results}
In our experiments we compare the results of with the Triplet loss baseline (trained on the same date with the same conditions) and also investigating the effect of self-morph presence.
We present our results as the MMPMR dependency from the FNMR values (see Fig. \ref{fig:MMPMR_1}) and values of MMPMR at the FNMR=0.01 (see Table \ref{tab:MMPMR_results}).

% \usepackage{multirow}
\begin{table}[htbp]
\caption{MinMax-MMPMR and ProdAvg-MMPMR for different models and protocols at FNMR=0.01.}
\centering
\begin{tabular}{|c|cc|}
\hline
\multirow{2}{*}{Model \& Protocol} & \multicolumn{2}{c|}{Rate @ FNMR=0.01} \\ \cline{2-3} 
                                   & \multicolumn{1}{c|}{MinMax-MMPMR}  & ProdAvg-MMPMR  \\ \hline
                            Triplet LDM    & \multicolumn{1}{c|}{0.99953}         &   0.97890    \\ \hline
                            Triplet+Self LDM       & \multicolumn{1}{c|}{1.0}         &   0.97849   \\ \hline
                            Triplet STD        & \multicolumn{1}{c|}{0.99906}         &    0.96828   \\ \hline
                            Triplet+Self STG        & \multicolumn{1}{c|}{1.0}         &   0.97456    \\ \hline
                            Quadruplet LDM      & \multicolumn{1}{c|}{0.99766}         &    0.93530   \\ \hline
                            Quadruplet+Self LDM       & \multicolumn{1}{c|}{1.0}         &    0.978870   \\ \hline
                            Quadruplet STG       & \multicolumn{1}{c|}{0.91876}         &   0.642203    \\ \hline
                            Quadruplet+Self STG       & \multicolumn{1}{c|}{0.999533}         &   0.9666627    \\ \hline
\end{tabular}
\label{tab:MMPMR_results}
\end{table}

Withing the same setup Quadruplet Loss with our sampling strategy allowed to achieve better face morphing robustness then the Triplet Loss baseline.  At the same time self morphs didn't demonstrate significant impact to the performance.

\section{Conclusion}

In this work we revisit the contrastive methods in face recognition and introduce a novel quadruplet loss function, which is designed to improve the robustness of face recognition systems against morphing attacks. Through the specific sampling of quadruplets of face images, coupled with face morphs, our approach demonstrates promising results in enhancing the resilience of face recognition networks against morphing attacks. By introducing the formulation of the quadruplet loss and conducting extensive experiments across various data settings, we prove the effectiveness of our strategy in resisting face morphing attacks. In our future work we plan to extend this study by enlarging our benchmark utilities with other types of morph samples and also print/scan morphs and perform corresponding experiments. We also plan to perform a extensive study of our benchmark utilities with various face recognition utilities and networks.

%%%%%%%%% REFERENCES
{\small
\bibliographystyle{IEEEtran}
\bibliography{egbib}

}

\end{document}